\documentclass{article}
\usepackage[utf8]{inputenc}
\usepackage{amsmath}
\usepackage{amssymb}
\usepackage{amsthm}
\usepackage{mathtools}
\usepackage{etex}
\usepackage{etoolbox}
\usepackage{xparse}
\usepackage{caption}
\usepackage{subcaption}

\usepackage{microtype}
\usepackage{graphicx}
\usepackage{tikz}
\usepackage{booktabs} 

\usepackage[export]{adjustbox}[2011/08/13]

\usepackage[toc,nopostdot]{glossaries}

\usepackage[accepted]{icml2021}

\usepackage[hidelinks]{hyperref}

\usepackage{cleveref}

\makeatletter

\renewenvironment{proof}[1][\proofname]{\par
  \pushQED{\qed}%
  \normalfont \partopsep=\z@skip \topsep=\z@skip
  \trivlist
  \item[\hskip\labelsep
        \itshape
    #1\@addpunct{.}]\ignorespaces
}{%
  \popQED\endtrivlist\@endpefalse
}

\allowdisplaybreaks

\graphicspath{{figures/}}

\newcommand{\argmax}{\operatornamewithlimits{argmax}}
\newcommand{\argmin}{\operatornamewithlimits{argmin}}

\definecolor{i6blue}{rgb}{0.0, 0.4, 0.8}
\definecolor{i6deepblue}{rgb}{0.0, 0.2, 1.0}
\definecolor{i6green}{rgb}{0.4, 0.8, 0.0}
\definecolor{i6lightblue}{rgb}{0.6, 0.9, 1.0}
\definecolor{i6orange}{rgb}{0.8, 0.4, 0.0}
\definecolor{i6pink}{rgb}{0.8, 0.0, 0.4}

\definecolor{orange}{rgb}{1.0, 0.6, 0.0}
\definecolor{yellow}{rgb}{1.0, 1.0, 0.0}

\definecolor{grey}{rgb}{0.6, 0.6, 0.6}
\definecolor{lightgrey}{rgb}{0.9, 0.9, 0.9}

\definecolor{i6bluedark}{rgb}{0.0156,0.2578,0.5625}

\definecolor{deepblue}{rgb}{0,0,0.5}
\definecolor{deepred}{rgb}{0.6,0,0}
\definecolor{deepgreen}{rgb}{0,0.4,0}
\definecolor{gray}{rgb}{0.5,0.5,0.5}

\definecolor{mygreen}{rgb}{0,0.6,0}
\definecolor{mygray}{rgb}{0.5,0.5,0.5}
\definecolor{mymauve}{rgb}{0.58,0,0.82}

\definecolor{delim}{RGB}{20,105,176}

\def\isechsblue#1    {\textcolor{i6blue} {#1} }
\def\isechsbluedark#1    {\textcolor{i6bluedark} {#1} }
\def\blue#1    {\textcolor{blue} {#1} }
\def\green#1   {\textcolor{green} {#1} }
\def\magenta#1 {\textcolor{magenta} {#1} }

\newsavebox\CBox

\DeclareRobustCommand\onedot{\futurelet\@let@token\@onedot}
\def\@onedot{\ifx\@let@token.\else.\null\fi\xspace}

\def\R{\ensuremath{\mathbb{R}}}
\def\N{\ensuremath{\mathbb{N}}}

\def\E{\ensuremath{\mathbb{E}}}

\newcommand{\softmax}{\operatorname{softmax}}

\def\mid@vertical{\mskip1mu\vrule\mskip1mu}
\def\midvert{\egroup\;\mid@vertical\;\bgroup}
\NewDocumentCommand\Set{mg}{%
    \IfNoValueTF{#2}{%
        \ensuremath{\left\{ #1 \right\}}%
    }{%
        \ensuremath{\left\{ {#1} \;\mid@vertical\; {#2} \right\}}%
    }%
}

\newcommand{\appropto}{\mathrel{\vcenter{
  \offinterlineskip\halign{\hfil$##$\cr
    \propto\cr\noalign{\kern2pt}\sim\cr\noalign{\kern-2pt}}}}}

\newcommand\numberthis{\addtocounter{equation}{1}\tag{\theequation}}

\newcommand{\crnn}{\gls{returnn}}

\newcommand{\charsym}[1]{\textnormal{\texttt{#1}}}
\newcommand{\charblank}{\charsym{B}}
\newcommand{\charsyma}{\charsym{a}}
\newcommand{\charsymb}{\charsym{b}}
\newcommand{\charsymc}{\charsym{c}}
\newcommand{\genblank}{\mathring{s}}
\newcommand{\blank}{\operatorname{blank}}

\newcommand{\sil}{\textrm{silence}}

\theoremstyle{plain}\newtheorem{lemma}{Lemma}[section]
\theoremstyle{plain}\newtheorem{theorem}[lemma]{Theorem}
\theoremstyle{plain}
\theoremstyle{plain}\newtheorem{corollary}[lemma]{Corollary}
\theoremstyle{definition}\newtheorem{mydef}[lemma]{Definition}
\theoremstyle{definition}
\theoremstyle{definition}\newtheorem{example}[lemma]{Example}
\theoremstyle{definition}\newtheorem{remark}[lemma]{Remark}
\theoremstyle{definition}\newtheorem{prelim}[lemma]{Preliminaries}
\theoremstyle{definition}\newtheorem{simulation}[lemma]{Simulation}

\newcommand{\ExternalLink}{%
    \tikz[x=1.2ex, y=1.2ex, baseline=-0.05ex]{%
        \begin{scope}[x=1ex, y=1ex]
            \clip (-0.1,-0.1) 
                --++ (-0, 1.2) 
                --++ (0.6, 0) 
                --++ (0, -0.6) 
                --++ (0.6, 0) 
                --++ (0, -1);
            \path[draw, 
                line width = 0.5, 
                rounded corners=0.5] 
                (0,0) rectangle (1,1);
        \end{scope}
        \path[draw, line width = 0.5] (0.5, 0.5) 
            -- (1, 1);
        \path[draw, line width = 0.5] (0.6, 1) 
            -- (1, 1) -- (1, 0.6);
        }}

\makeatletter
\newcommand*{\bbldefcitehref}[2]{%
  \immediate\write\@mainaux{\string\auxdefcitehref{\unexpanded{#1}}{\unexpanded{#2}}}}

\newcommand*{\auxdefcitehref}[2]{%
  \expandafter\global\expandafter\def\csname citehref@#1\endcsname{#2}}

\newcommand*{\mkcitehref}[1]{%
  \ifcsname citehref@#1\endcsname
    \ \href{\csname citehref@#1\endcsname}{\ExternalLink}%
  \fi}

\def\NAT@hyper@#1{%
 \hyper@natlinkstart{\@citeb\@extra@b@citeb}#1\hyper@natlinkend%
 {\mkcitehref{\@citeb}}%
}

\makeatother

\glsdisablehyper

\newacronym[
description={\Glsentrylong{asr}. The conversion from speech-to-text by automatic means.}
]{asr}{ASR}{automatic speech recognition}

\newacronym[description={\Glsentrylong{lvcsr}.}
]{lvcsr}{LVCSR}{large vocabulary continuous speech recognition}

\newacronym[
description={\Glsentrylong{wer}. Error measure commonly used in \gls{asr}. Computes the Levenshtein distance \cite{levenshtein66} on a hypothesis and its reference.}
]{wer}{WER}{word error rate}

\newacronym[
description={\Glsentrylong{fer}. Error measure commonly used in \gls{asr}; useful in general as an error measure for \gls{nn} models trained with some frame-wise criteria.}
]{fer}{FER}{frame error rate}

\newacronym[
description={\Glsentrylong{bpe} \cite{sennrich16bpe}. Subword units, usually based on characters.
See \Cref{sec:seq2seq:label-units} for an overview of label units.}
]{bpe}{BPE}{byte-pair encoding}

\newacronym[description={\Glsentrylong{am}.}]{am}{AM}{acoustic model}
\newacronym[description={\Glsentrylong{lm}.}]{lm}{LM}{language model}

\newacronym[
description={\Glsentrylong{hmm}. Statistical sequence-to-sequence model.
A system being modeled hereby is assumed to be a Markov process with unobserved states.
The emission probabilities are usually modeled by \Glspl{gmm} if this is not a \gls{hybrid-hmm}.
See \Cref{sec:seq2seq:hmm}.}
]{hmm}{HMM}{hidden Markov model}

\newacronym[
description={\Glsentrylong{gmm}. Statistical model.
For \gls{asr}, often used as part of a \gls{hmm}.}
]{gmm}{GMM}{Gaussian mixture model}

\newacronym[
description={\Glsentrylong{nn}. Statistical model.
See \Cref{chap:neural}.}
]{nn}{NN}{neural network}

\newacronym[
description={\Glsentrylong{ffnn}. \Glsentrylong{nn} model only using feed-forward (non-recurrent) connections.
See \Cref{sec:ffnn}.}
]{ffnn}{FFNN}{feed-forward neural network}

\newacronym[
description={\Glsentrylong{mlp}. Alternative term for \gls{ffnn}.},
]{mlp}{MLP}{multilayer perceptron}
\glsunset{mlp}  

\newacronym[
description={\Glsentrylong{cnn}. \Glsentrylong{nn} model using convolutions. See \Cref{sec:cnn}.}
]{cnn}{CNN}{convolutional neural network}

\newacronym[
description={\Glsentrylong{tdnn}. A special type of \gls{cnn}.}
]{tdnn}{TDNN}{time delay neural network}

\newacronym[
description={\Glsentrylong{rnn}. \Glsentrylong{nn} model using recurrent connections.
The \gls{lstm} is the most common RNN type.}
]{rnn}{RNN}{recurrent neural network}

\newacronym[
description={\Glsentrylong{lstm} \cite{hochreiter1997lstm}. \Glsentrylong{rnn} model with gating.
See \Cref{sec:lstm}.}
]{lstm}{LSTM}{long short-term memory}

\newacronym[description={Bidirectional \gls{lstm}. See \Cref{sec:lstm}.}
]{blstm}{BLSTM}{bidirectional \glsentryshort{lstm}}

\newacronym[
description={\Glsentrylong{rnnt} \cite{graves2012seqtransduction}.
Sequence-to-sequence model based on a \gls{rnn}.
See \Cref{sec:transducer}.}
]{rnnt}{RNN-T}{recurrent neural network transducer}

\newacronym[
description={\Glsentrylong{rna} \cite{sak2017neuralaligner}.
Sequence-to-sequence model based on a \gls{rnn}.
See \Cref{sec:transducer}.}
]{rna}{RNA}{recurrent neural aligner}

\newacronym[
description={\Glsentrylong{hat} \cite{variani2020hat}.
Sequence-to-sequence model based on a \gls{rnn}.
See \Cref{sec:transducer}.}
]{hat}{HAT}{hybrid autoregressive transducer}

\newacronym[
description={\Glsentrylong{lace} \cite{yu2016lace} is a type of \gls{cnn} used for \gls{asr}.
See \Cref{sec:cnn}.}
]{lace}{LACE}{layer-wise context expansion with attention}

\newacronym[
description={\Glsentrylong{sgd}. Optimization method usually used to train neural networks.
See \Cref{chap:training}.}
]{sgd}{SGD}{stochastic gradient descent}

\newacronym[
description={\Glsentrylong{ce}. Measure between two probability distributions, used as a loss to train neural networks.
See \Cref{sec:training:ce}.}
]{ce}{CE}{cross entropy}

\newacronym[
description={\Glsentrylong{ctc} \cite{graves2006connectionist}.
Type of neural network output,
label topology including the special $\blank$ symbol,
and also a sequence training loss.
See \Cref{sec:fullsum:ctc}.}
]{ctc}{CTC}{connectionist temporal classification}

\newacronym[
description={\Glsentrylong{mmi}. Discriminative sequence training criteria. See \Cref{sec:seqtrain}.}
]{mmi}{MMI}{maximum mutual information}

\newacronym[
description={\Glsentrylong{smbr}. Discriminative sequence training criteria. See \Cref{sec:seqtrain}.}
]{smbr}{sMBR}{state-level minimum Bayes risk}

\newacronym[
description={\Glsentrylong{mpe}. Discriminative sequence training criteria. See \Cref{sec:seqtrain}.}
]{mpe}{MPE}{minimum phone error}

\newglossaryentry{hybrid-hmm}{
name={Hybrid \gls{nn}-\gls{hmm}},
text={hybrid \gls{nn}-\gls{hmm}},
description={Statistical sequence-to-sequence model.
The emission probability in the \gls{hmm} is modeled by a \gls{nn}.
See \Cref{sec:seq2seq:hybrid-hmm}.}
}

\newacronym[
description={\Glsentrylong{cart} \cite{breiman1984cart}.
Commonly used for state-tying of context-dependent phonemes for \glspl{hybrid-hmm}.}
]{cart}{CART}{classification and regression tree}

\newacronym[description={\Glsentrylong{wfsa}.}]{wfsa}{WFSA}{weighted finite state automaton}
\newacronym[description={\Glsentrylong{wfst}.}]{wfst}{WFST}{weighted finite state transducer}

\newacronym[description={\Glsentrylong{mfcc}.}]{mfcc}{MFCC}{mel-frequency cepstral coefficient}

\newacronym[description={\Glsentrylong{vtln}. Normalization, used for audio features.}
]{vtln}{VTLN}{vocal tract length normalization}

\newacronym[description={\Glsentrylong{cv}.}]{cv}{CV}{cross validation}

\newacronym[description={\Glsentrylong{relu}. Activation function for \glspl{nn}.}
]{relu}{ReLU}{rectified linear unit}

\newacronym[
description={RWTH extensible training framework for universal recurrent neural networks.
Neural network framework developed partly in this thesis.
See \Cref{sec:software:returnn} and \Cref{chap:returnn}.}
]{returnn}{RETURNN}{RWTH extensible training framework for universal recurrent neural networks}
\glsunset{returnn}  

\newacronym[description={\Glsentrylong{cpu}.}]{cpu}{CPU}{central processing unit}
\glsunset{cpu}
\newacronym[description={\Glsentrylong{gpu}.}]{gpu}{GPU}{graphics processing unit}
\glsunset{gpu}
\newacronym[description={\Glsentrylong{tpu}.}]{tpu}{TPU}{tensor processing unit}
\glsunset{tpu}

\icmltitlerunning{Why does CTC result in peaky behavior?}

\begin{document}

\twocolumn[
\icmltitle{Why does CTC result in peaky behavior?}



\icmlsetsymbol{equal}{*}

\begin{icmlauthorlist}
\icmlauthor{Albert Zeyer}{rwth,apptek}
\icmlauthor{Ralf Schlüter}{rwth,apptek}
\icmlauthor{Hermann Ney}{rwth,apptek}
\end{icmlauthorlist}

\icmlaffiliation{rwth}{Human Language Technology and Pattern Recognition, Computer Science Department, RWTH Aachen University, Aachen, Germany}
\icmlaffiliation{apptek}{AppTek GmbH, Aachen, Germany}

\icmlcorrespondingauthor{Albert Zeyer}{zeyer@cs.rwth-aachen.de}

\icmlkeywords{CTC, peaky behavior, spiky behavior, HMM, formal analysis}

\vskip 0.3in
]



\printAffiliationsAndNotice{}  








% ...

\begin{abstract}
The peaky behavior of CTC models is well known experimentally.
However, an understanding about
\emph{why} peaky behavior occurs is missing,
and whether this is a good property.
We provide a formal analysis
of the peaky behavior and gradient descent convergence properties
of the CTC loss and related training criteria.
Our analysis provides a deep understanding why peaky behavior occurs
and when it is suboptimal.
On a simple example which should be trivial to learn for any model,
we prove that a feed-forward neural network
trained with CTC from uniform initialization
converges towards peaky behavior with a 100\% error rate.
Our analysis further explains
why CTC only works well together with the $\blank$ label.
We further demonstrate that peaky behavior does not occur
on other related losses including a label prior model,
and that this improves convergence.
\end{abstract}

\section{Introduction}

The peaky behavior of \gls{ctc} \cite{graves2006connectionist} (\Cref{fig:ctc-full-sum-peaky-out-2})
was often observed experimentally.
However, it is not well explained and analyzed
\emph{why} models trained with \gls{ctc} get peaky.
Also, other training criteria for the same models and label sets
will not result in the same peaky behavior,
so this is a result of the \gls{ctc} training criterion.

We will formally define our understanding of peaky behavior.
Then we provide a formal analysis as to
in what cases and \emph{why} we will get such behavior.
We will see that the peaky behavior results as a corollary from the training criterion
and its local convergence properties,
where gradient descent from a uniform initialization
tends towards suboptimal local optima with peaky behavior.

We demonstrate that in the case of a training criterion with peaky behavior like \gls{ctc},
it is crucial to use a label topology with a $\blank$ label,
and having a silence label in case of speech recognition is suboptimal.
This is an important new understanding
of the $\blank$ label,
which has been observed experimentally before
\cite{bluche-blog-ctc,bluche2015ctc,bluche2015phd}.

Peaky behavior can be problematic in certain cases,
e.g.~when an application requires to not use the $\blank$ label,
e.g.~to get meaningful time accurate alignments of phonemes to a transcription.
Also, we will mathematically demonstrate
that local-context models like \glspl{ffnn} are suboptimal
for such kind of training criterion like \gls{ctc},
which is due to the peaky behavior.
%
We show variations of the training criterion by including a label prior,
and we demonstrate that this solves convergence problems
and does not lead to peaky behavior anymore.


Some of the mathematical proofs and demonstrations were assisted using
the computer algebra system SymPy \cite{sympy2017}
via symbolic computation.
We also performed synthetic experiments using
TensorFlow \cite{tensorflow2015}
and \crnn{} \cite{zeyer2018:returnn}. 
We publish all the symbolic computation code,
and all the code and configs of our experiments.
\footnote{\tiny\url{https://github.com/rwth-i6/returnn-experiments/tree/master/2021-formal-peaky-behavior-ctc}}

\section{Related Work}



While this work focuses on \gls{ctc}, 
we have a similar complete marginalization over all possible alignments
in the training criterion of \gls{rnnt}
\cite{graves2012seqtransduction},
\gls{rna} \cite{sak2017neuralaligner},
lattice-free \gls{mmi} \cite{povey2016purely}  
and AutoSegCriterion (ASG) \cite{collobert2016wav2letter}.

Training with the full-sum over all alignment paths with neural networks
is not novel
\cite{
bengio1991nnhmm,
haffner1993connectionistmmi,
senior1996forward,
hennebert1997estimation,
yan1997speech,
lecun1998gradient,
li2014labelingunsegmented,
bluche2015ctc,
zeyer2017:ctc}.
From-scratch (flat-start, \gls{gmm}-free)
training with frequent realignments was also discussed in \cite{zhang2014standalone,senior2014gmmfree,bacchiani2014asynchronous}.

We see later the importance and impact of the label prior model
on peaky behavior and convergence behavior.
A label prior model has been used together with CTC-trained models
at decoding time \cite{kanda2016mapctc,miao2015eesen}
but only rarely at training time \cite{zeyer2017:ctc}.

Some attempts to explain the peaky behavior can be found in
\cite{bluche-blog-ctc,bluche2015ctc,bluche2015phd}.
The work addresses many related questions
such as peaky behavior, convergence behavior
and the role of the blank label.
To the best of our knowledge,
no prior work exists which formally analyzes the reasons of peaky behavior.
It is not well explained
why such training criteria results in this unnatural behavior,
and in fact other training criteria do not.

\section{Definition of Peaky Behavior}

\begin{figure}
\centering
\includegraphics[width=1.\linewidth]{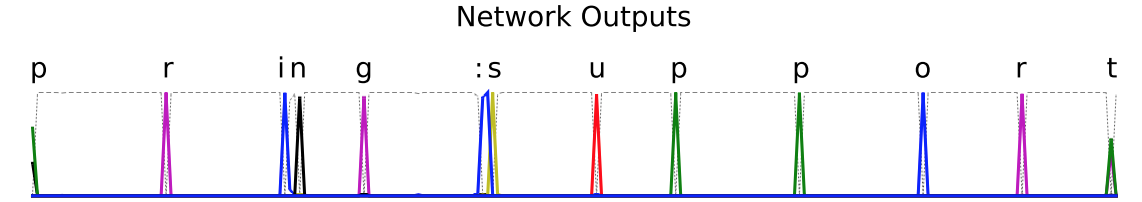}
\vspace{-6mm}
\caption{\Gls{ctc} peaky output, Figure 7.9 from \cite{graves2012sequencelabel}.
The colored lines depict the output activations for different labels over time.
The greyed dotted line represents the blank label.}
\label{fig:ctc-full-sum-peaky-out-2}
\end{figure}

The peaky behavior of \gls{ctc} is best illustrated in
\Cref{fig:ctc-full-sum-peaky-out-2}.
It shows the \gls{nn} output probability distribution over a subset of the labels,
including the $\blank$ label.
It can be seen that along the time axis,
the $\blank$ label is dominating most of the time,
while the other labels are observed only as a spike event.
I.e.\ the probability distribution over the time is peaky.

\begin{prelim}
Let $S$ be a finite set of labels.
Let $\mathcal{M}$ be a model such that it defines
the probability distribution
$p_t(s | x_1^T, \mathcal{M})$
for some input signal $x_1^T$,
over time frames $t \in \Set{1, \dots, T}$, $s \in S$.
For notational simplicity, we partially leave out the condition on $\mathcal{M}$.
The probability of the label sequence $y_1^N$
is defined assuming label-independence
\[ p(y_1^N|x_1^T) = \sum_{s_1^T : y_1^N} p(s_1^T|x_1^T)
= \sum_{s_1^T : y_1^N} \prod_t p_t(s_t | x_1^T) . \]
We call $s_1^T \in S^T$ an alignment.
The elements of $y$ are not relevant here --
what matters are all the allowed alignments $s_1^T$ given $y_1^N$.
We denote the set of possible alignments as
$\mathcal{A}_T(y_1^N) := \Set{s_1^T}{s_1^T : y_1^N}$.
$\mathcal{A}$ is also called the \emph{label topology}
and is usually defined by a finite state transducer (FST).
In our case, we always use a FST which is equivalent
to a regular expression (RE)
of the form $y_1^{1|+|*} \dots y_n ^{1|+|*}$, $y \in S$.
For simplicity, we assume this defines a unique mapping $(T, s_1^T) \mapsto (N, y_1^N)$.
Let $\mathcal{D} = \Set{(x_1^T,y_1^N)}$ be the training dataset.
In the following, we focus the analysis only on a single training sample $(x_1^T,y_1^N)$.
This is not really a restriction, as you can concatenate multiple sequences into one.
Also, in most cases, the analysis would trivially generalize to multiple sequences
but would complicate the notation.
\end{prelim}

The \emph{label topology} $\mathcal{A}$ will be of central importance
for the convergence behavior.
CTC defines such a topology by allowing $\blank$ anywhere,
and by allowing label repetitions.
We explicitly define some possible topologies
for some simple examples.

\newcommand{\exre}{\charblank^* \charsyma^+ \charblank^*}
\begin{example}\label{def:fullsum:example1}
Consider the target sequence consisting only of a single label,
$y_1^N = (\charsyma)$, and define our label set as $S = \Set{\charblank,\charsyma}$,
and
\[ \mathcal{A}_T(y_1^N) := \Set{s_1^T}{\text{$s_1^T$ matches reg.~expr.~$\exre$}} . \]
\end{example}
Despite an empty target sequence,
\Cref{def:fullsum:example1} is arguably one of the simplest possible non-trivial \gls{ctc} examples.
The label $\charblank$ plays the role of the $\blank$ label.


\begin{mydef}[Peaky alignment]
Let $s_1^T \in \mathcal{A}_T(y_1^N)$.
We say that the \emph{alignment} $s_1^T$ is \emph{peaky with dominant label $\genblank$},
if
\[ \left|\Set{t}{s_t = \genblank} \right| > \left|\Set{t}{{\tilde{s}}_t = \genblank}\right|
\quad \forall\ {\tilde{s}}_1^T \in \mathcal{A}_T(y_1^N) . \]
\end{mydef}

With \Cref{def:fullsum:example1}, and $T=100$,
$s_1^T = (\charblank^{49}, \charsyma, \charblank^{50})$
is a peaky alignment with dominant label $\genblank = \charblank$.
Label $\charsyma$ occurs only in one single frame, i.e.~is peaky.
I.e.~a peaky alignment is peaky w.r.t.~all the non-dominant labels.

\begin{mydef}[Viterbi]
Given a model $\mathcal{M}$ and a sample $(x_1^T,y_1^N)$,
a \emph{Viterbi} alignment is an alignment $s_1^T \in \mathcal{A}_T(y_1^N)$
such that it maximizes $\prod_t p_t(s_t | x_1^T, \mathcal{M})$.
\end{mydef}

\begin{mydef}[Peaky behavior]
Given a sample $(x_1^T,y_1^N)$ and a model $\mathcal{M}$.
If all Viterbi alignments $s_1^T \in \mathcal{A}_T(y_1^N)$ are \emph{peaky},
then the \emph{model} $\mathcal{M}$ has
\emph{peaky behavior for $(x_1^T,y_1^N)$}.
If that holds true for all $(x_1^T, y_1^N) \in \mathcal{D}$,
then we simply say that the model has \emph{peaky behavior}.
\end{mydef}


\begin{mydef}[Alignment count]
Let
\[ \mathcal{C}(s, t, T)
:= \left| \Set{s_1^T}{s_t = s,  s_1^T \in \mathcal{A}_T(y_1^N)} \right| \]
be the \emph{count} of all alignments with label $s$ in frame $t$.
Let
\[ \mathcal{C}(T) := \left| \mathcal{A}_T(y_1^N) \right| \]
be the total \emph{count} of all alignments. 
\end{mydef}

\begin{remark}
Note that the alignment count is totally independent
from the input features $x$. 
It only depends on the possible alignments in the label topology $\mathcal{A}_T(y_1^N)$.
\end{remark}

\begin{mydef}[Label count]
Let
\[ \mathcal{C}(s,T) := \sum_t \mathcal{C}(s, t, T) \]
be the total \emph{count} of label $s \in S$ in all frames.
\end{mydef}

\begin{mydef}[Dominant label]
The label $\genblank \in S$ is \emph{dominant in $\mathcal{A}_T(y_1^N)$},
if
\[ \mathcal{C}(\genblank) > \mathcal{C}(s)
\quad \forall\ s \in S, s \neq \genblank . \]
\end{mydef}

\begin{remark}
\label{rem:fullsum:dom-label-by-topo}
Note that the dominance property of $\genblank$ is defined
depending on $T, y_1^N$ and the label topology $\mathcal{A}$,
\emph{independent from the input $x$}.
For the \gls{ctc} topology, $\blank$ always has this property.
For the common \gls{hmm} topology in speech recognition,
$\sil$ almost always gets this property,
simply by the same counting arguments.
\end{remark}

Let us recall the \Cref{def:fullsum:example1} ($\exre$).
This example is simple enough that we can exactly calculate these counts.

\begin{lemma}\label{lem:fullsum:ex1:count}
For \Cref{def:fullsum:example1} ($\exre$),
\begin{align*}
\mathcal{C}(T) &= \frac{T \cdot (T + 1)}{2} \\
\mathcal{C}(s{=}\charsyma, t, T) &= t \cdot (T - t + 1) \\
\mathcal{C}(s{=}\charblank, t, T) &= \frac{T^2}{2} - T\cdot t + \frac{T}{2} + t^2 - t \\
\mathcal{C}(s{=}\charsyma, T) &= \frac{T \cdot (T^2 + 3 T + 2)}{6} \\
\mathcal{C}(s{=}\charblank, T) &= \frac{T \cdot (T^2 - 1)}{3}
\end{align*}
Thus the \emph{dominant label is $\genblank = \charblank$} for $T \ge 5$.
\end{lemma}

\begin{corollary}\label{cor:fullsum:ex1:avg-dom-label}
Following from \Cref{def:fullsum:example1} ($\exre$) and \Cref{lem:fullsum:ex1:count}
we get
\[ \frac{\mathcal{C}(\genblank, T)}{\sum_s \mathcal{C}(s, T)}
= \frac{2 \cdot (T - 1)}{3 T} \]
which is the average count of the dominant label $\genblank = \charblank$ per frame.
I.e.~for $T \ge 5$ we have
$\frac{\mathcal{C}(\genblank, T)}{\sum_s \mathcal{C}(s, T)} > 50\%$.
\end{corollary}

\begin{corollary}\label{cor:fullsum:ex1:countblankframes}
We can count the number of frames where some label $s$ dominates,
i.e.~define
\[ C^{\text{Frames}}_T(s) := \left| \Set{t}{\mathcal{C}(s,t,T) > \mathcal{C}(s',t,T) \ \forall s' \ne s} \right| \le T . \]
%
Following further from \Cref{def:fullsum:example1} ($\exre$) and \Cref{lem:fullsum:ex1:count}
we get
\begin{align*}
C^{\text{Frames}}_T(\genblank) &= 2 \lceil  \tfrac{1}{2} T -  \tfrac{1}{2} \sqrt{T + 1} -  \tfrac{1}{2} \rceil
\ge T - \sqrt{T + 1} - 1 .
\end{align*}
I.e.~$\lim_{T\rightarrow \infty} \frac{C^{\text{Frames}}_T(\genblank)}{T} = 1$,
i.e.~in the limit, $\genblank = \charblank$ will strongly dominate per frame.
For $T \ge 8$, we have $\frac{C^{\text{Frames}}_T(\genblank)}{T} \ge 50\%$.
For $T \ge 24$, we have $\frac{C^{\text{Frames}}_T(\genblank)}{T} \ge 75\%$.
\end{corollary}
Recall again that these corollaries
are just about the dataset $\mathcal{D}$,
or more precisely just the target label sequences $y_1^N$
and the input sequence length $T$ (but not the input itself, $x_1^T$),
and the topology $\mathcal{A}$.
They are independent from any training criterion or any model.
However, based on these, we will show that models trained with the \gls{ctc} criterion
with gradient descent show peaky behavior,
i.e.~all their Viterbi alignments are peaky.


\section{Convergence to Peaky Behavior}

Now we study the convergence behavior of the training criterion (loss)
when trained with gradient descent.
I.e.~we have some model initialization and we locally modify the model parameters
such that the loss decreases.


\begin{mydef}
\label{def:fullsum:loss}
The CTC loss is defined as
\[ L :=
- \log \sum_{s_1^T : y_1^N} p(s_1^T | x_1^T)
=
- \log \sum_{s_1^T : y_1^N} \prod_t p_t(s_t | x_1^T) . \]
\end{mydef}

\begin{remark}\label{rem:fullsum:sharp-globalopt}
We always have $L \ge 0$.
Let $s_1^T$ be any valid alignment (peaky or not),
and assume a model $\hat{\mathcal{M}} (s_1^T)$
with
\[ p_t(s | x_1^T, \hat{\mathcal{M}} (s_1^T))
:= \begin{cases} 1 &\text{if } s = s_t , \\ 0 &\text{else} . \end{cases} . \]
Then we have reached a global optimum with $L=0$.
If any $p_t$ is not sharp like this, we have $L > 0$.
\end{remark}

\begin{remark}\label{def:fullsum:q}
Let $\theta$ be the model parameters of $\mathcal{M}$.
The gradient of $L$ with respect to the model parameters $\theta$
is given (compare \cite{graves2012sequencelabel,zeyer2017:ctc}) as
\begin{align}
\label{eq:fullsum:L-grad}
\frac{\partial}{\partial \theta} L
& =
-\sum_{s,t} q_t(s | x_1^T, y_1^N, \theta)
\cdot \frac{\partial}{\partial \theta} \log p_t(s_t | x_1^T, \theta)
\end{align}
with
\begin{align*}
q_t(s | x_1^T, y_1^N, \theta) =
\frac{
\sum_{s_1^T : y_1^N, s_t=s}
p(s_1^T | x_1^T, \theta)}{
\sum_{s_1^T : y_1^N}
p(s_1^T | x_1^T, \theta)}
.
\end{align*}
The quantity $q$ 
can be efficiently computed using the forward-backward (Baum-Welch)
algorithm
and is also known as soft-alignment.
\end{remark}
\begin{remark}\label{rem:q-by-counting}
If $p_t$ is a uniform distribution for all $t$,
it cancels out in $L$ and also in $q_t$.
We simply get
\begin{align*}
q_t(s | x_1^T, y_1^N, \theta)
= \frac{\sum_{s_1^T : y_1^N, s_t=s} 1}{\sum_{s_1^T : y_1^N} 1}
= \frac{\mathcal{C}(s, t, T)}{
\mathcal{C}(T)} .
\end{align*}
\end{remark}

As a first model to understand the convergence behavior,
we analyze a model which is totally independent from the input $x_1^T$
and just consists of a bias term.
We would expect that this model learns a prior over the labels
as they occur in the training targets.
The model is also relevant,
as every neural network usually has a bias term in the output softmax,
and this bias term will get exactly the same gradient. 

\begin{mydef}[Bias model]\label{def:fullsum:biasmodel}
The model $\mathcal{M}^b$
just consists of a single bias parameter,
and is totally independent from the input $x_1^T$,
i.e.~for $\theta = b \in \R^S$,
\[ p_t(s|x_1^T,\mathcal{M}^b) := \softmax(b) . \]
\end{mydef}

\begin{theorem}
\label{th:fullsum:biasmodel}
Let $\genblank$ be dominant in $\mathcal{A}_T(y_1^N)$.
Starting with the model $\mathcal{M}^{b_0}$ initialized with uniform distribution,
for example $b_0 = 0$,
then gradient descent on $L$ will converge to a model with peaky behavior.

\begin{proof}
First observe that
\begin{align*}
\frac{\partial}{\partial b_i} L
= \sum_t p_t - q_t
= T \cdot \left( \softmax(b_i) - \E_t [q_t] \right)
\end{align*}
for a gradient step $i$.
By \Cref{rem:q-by-counting} for $i=0$ we get
\begin{align*}
\mathbb{E}_t [q_t] [s]
%
%
%
= \frac{
\mathcal{C}(s, T)}{
\sum_{s' \in S} \mathcal{C}(s', T)}
.
\end{align*}
I.e.~$\argmin_s \frac{\partial}{\partial \theta_0} L = \genblank$.
One gradient step will result in $b_1[\genblank] > b_1[s] \quad\forall s \ne \genblank$.
For $|S| = 2$, it is clear that we cannot escape from that region of $b$ anymore
where we always have peaky behavior.
For the case $|S| > 2$, for some $s \ne \genblank$,
when comparing the relative difference of $q[\genblank]$ vs.~$q[s]$
in the forward-backward computation through the FST,
we can disregard any paths not contributing to $\{\genblank, s\}$,
as they will be shared.
Thus we can reduce the case to $|S| = 2$,
and it follows that $b[\genblank] > b[s]$.
\end{proof}
\end{theorem}

\begin{simulation}
%
Consider \Cref{def:fullsum:example1} ($\exre$), $T=5$,
with dominant label $\genblank = \charblank$ (via \Cref{lem:fullsum:ex1:count}).
We can simulate that
the bias model uniformly initialized converges to the probability distribution
$p_t(s|x_1^T) \approx \{\charblank \mapsto 0.72, \charsyma \mapsto 0.28 \}$ (i.e.~peaky behavior),
which does not reflect the target label prior distribution
$\frac{\mathcal{C}(s, T)}{\sum_{s'} \mathcal{C}(s', T)} \approx \{\charblank \mapsto 0.53, \charsyma \mapsto 0.47 \}$.
%
I.e.~\emph{peaky behavior reinforces itself}.
\end{simulation}


%
%

Now we consider a very simple model with dependence on the input $x$.
This can be interpret as a \gls{ffnn} with a single softmax layer and no bias.
\begin{mydef}[FFNN]\label{def:fullsum:ffnn}
Define the model $\mathcal{M}^W$ as
\[ p_t(s|x_1^T) := \softmax(x_t W)[s], \]
where $W \in \R^{D_x, S}$ and $x_t \in \R^{D_x}$.
\end{mydef}

\begin{example}\label{fullsum:example1:input}
For \Cref{def:fullsum:example1} ($\exre$), define
\[ x_1^T := \left(
\smash[b]{\underbrace{\begin{matrix}0 \\ 1\end{matrix} \cdots \begin{matrix}0 \\ 1\end{matrix}}_{\text{$n$ times}}}\quad
\smash[b]{\underbrace{\begin{matrix}1 \\ 0\end{matrix} \cdots \begin{matrix}1 \\ 0\end{matrix}}_{\text{$2n$ times}}}\quad
\smash[b]{\underbrace{\begin{matrix}0 \\ 1\end{matrix} \cdots \begin{matrix}0 \\ 1\end{matrix}}_{\text{$n$ times}}}
\right)
\vphantom{\underbrace{\begin{matrix}0 \\ 1\end{matrix} \cdots \begin{matrix}0 \\ 1\end{matrix}}_{\text{$n$ times}}} ,
\]
for some $n \in \N$, $T = 4n$,
i.e.~$x_t \in \R^2$ and either $x_t = x_\charblank := \left( \begin{smallmatrix}0 \\ 1\end{smallmatrix} \right)$
or $x_t = x_\charsyma := \left( \begin{smallmatrix}1 \\ 0\end{smallmatrix} \right)$.
These constructed $x_t$ can be interpreted as corresponding
to the label $\charblank$ or label $\charsyma$.
\end{example}

\begin{remark}
Note that \Cref{fullsum:example1:input} is constructed in such a way that the probability distribution $p(x)$
over possible inputs $x$ is uniform.
This is optimistic because in practice, e.g.~in audio, silence frames often dominate.
Such input domination contributes further to peaky behavior.
However, we will show that we get peaky behavior even for this constructed case
where there is no dominating input feature.
\end{remark}

\begin{remark}\label{fullsum:example1:ffnn:globalopt}
With \Cref{fullsum:example1:input} and the \gls{ffnn} model, we can see that we reach the optimum $L=0$
with $W = \left( \begin{smallmatrix} \infty & 0 \\ 0 & \infty \end{smallmatrix} \right)$
and get as close as we want with a matrix over $\R$.
Any such solution has 0\% error rate.
This trivially generalizes to similarly constructed more complex examples. 
\end{remark}

\begin{figure*}
\centering
\captionsetup[subfigure]{justification=centering}
\begin{subfigure}[t]{0.24\linewidth}
\centering
\includegraphics[width=1.3\linewidth,center]{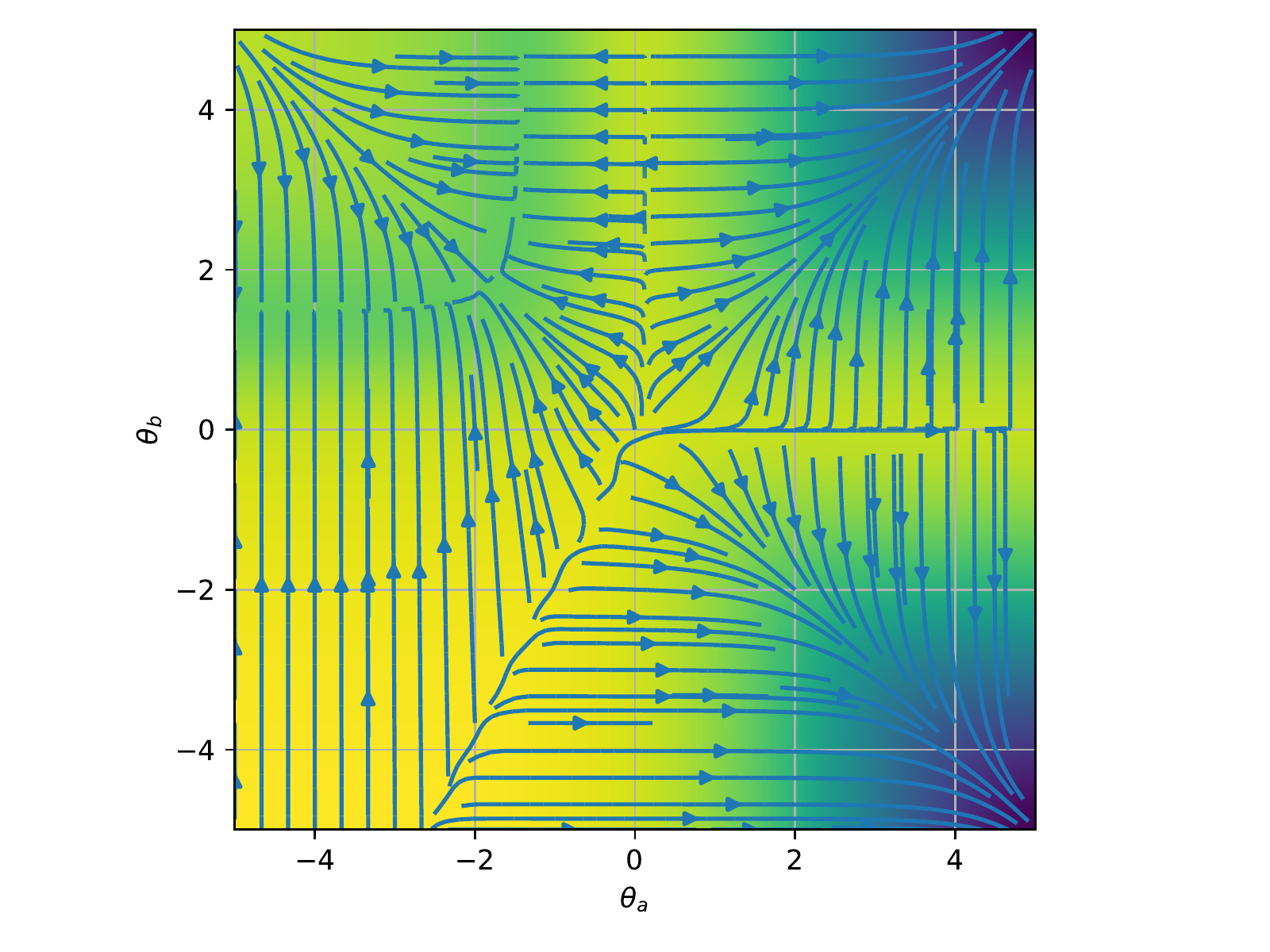}
\vspace{-6mm}
\caption{$L$ (CTC) for FFNN (\Cref{sec:fullsum:app:discriminative-sum}).%
}
\label{fig:fullsum:app:loss-map-discriminative-sum}
\end{subfigure}\hfill%
\begin{subfigure}[t]{0.24\linewidth}
\centering
\includegraphics[width=1.3\linewidth,center]{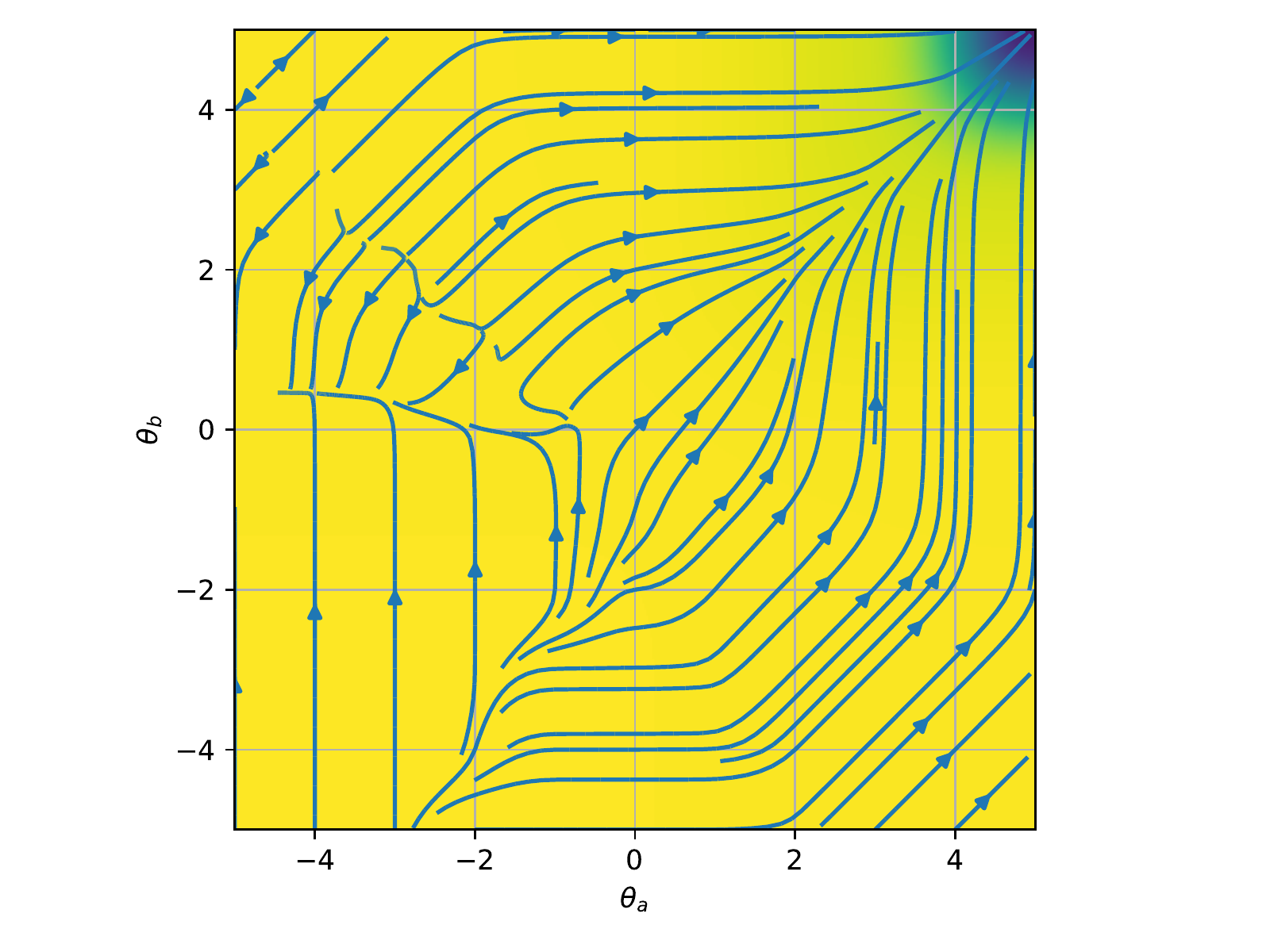}
\vspace{-6mm}
\caption{$L_{\text{hybrid}}$ for FFNN
(\Cref{sec:fullsum:app:discriminative-sum-with-prior}).%
}
\label{fig:fullsum:app:loss-map-discriminative-sum-with-prior}
\end{subfigure}\hfill%
\begin{subfigure}[t]{0.24\linewidth}
\centering
\includegraphics[width=1.3\linewidth,center]{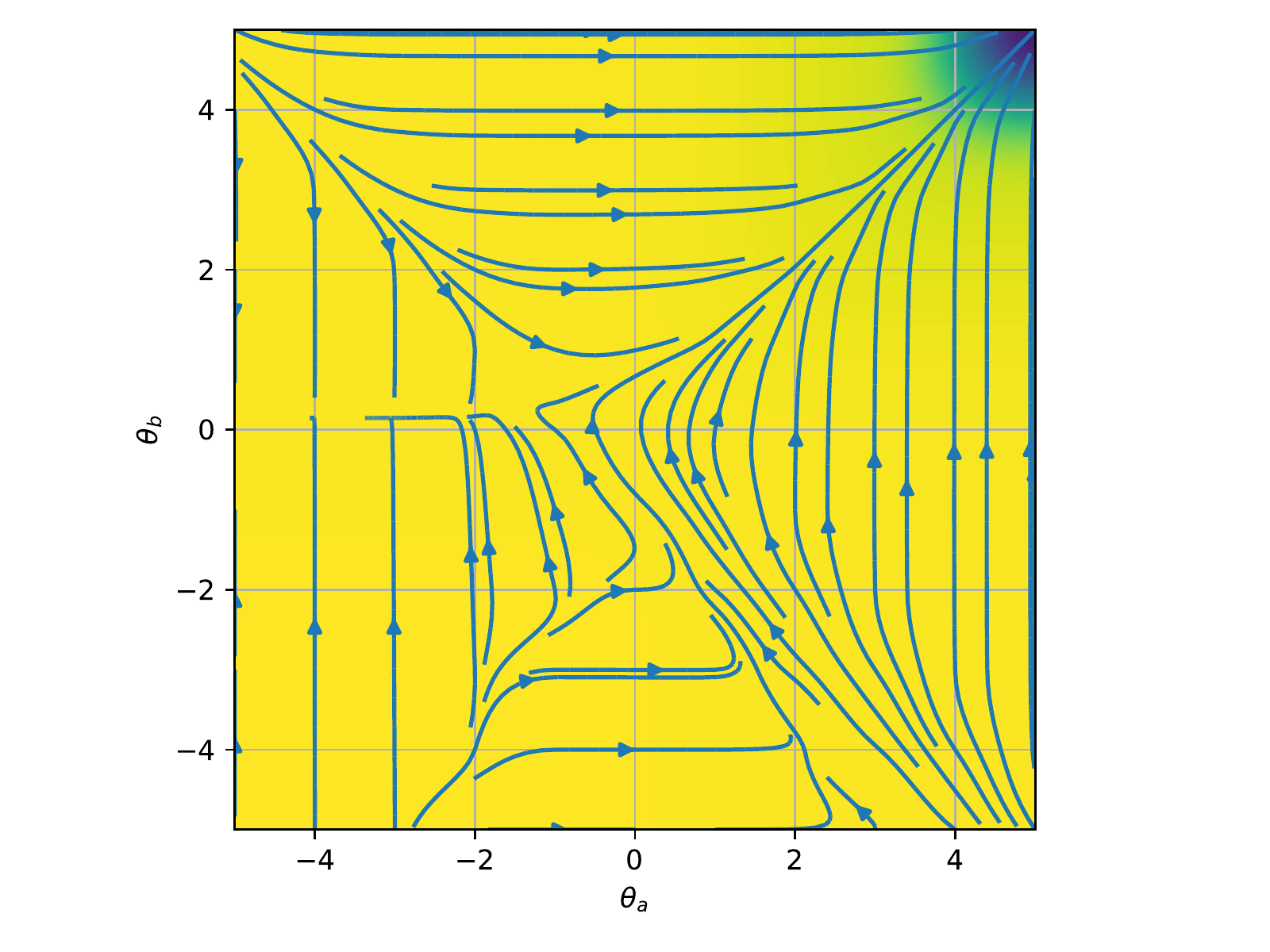}
\vspace{-6mm}
\caption{$L_{\text{hybrid}}$ for FFNN,
stop-grad.~on prior
(\Cref{sec:fullsum:app:discriminative-sum-with-prior-sg}).%
}
\label{fig:fullsum:app:loss-map-discriminative-sum-with-prior-sg}
\end{subfigure}\hfill%
\begin{subfigure}[t]{0.24\linewidth}
\centering
\includegraphics[width=1.3\linewidth,center]{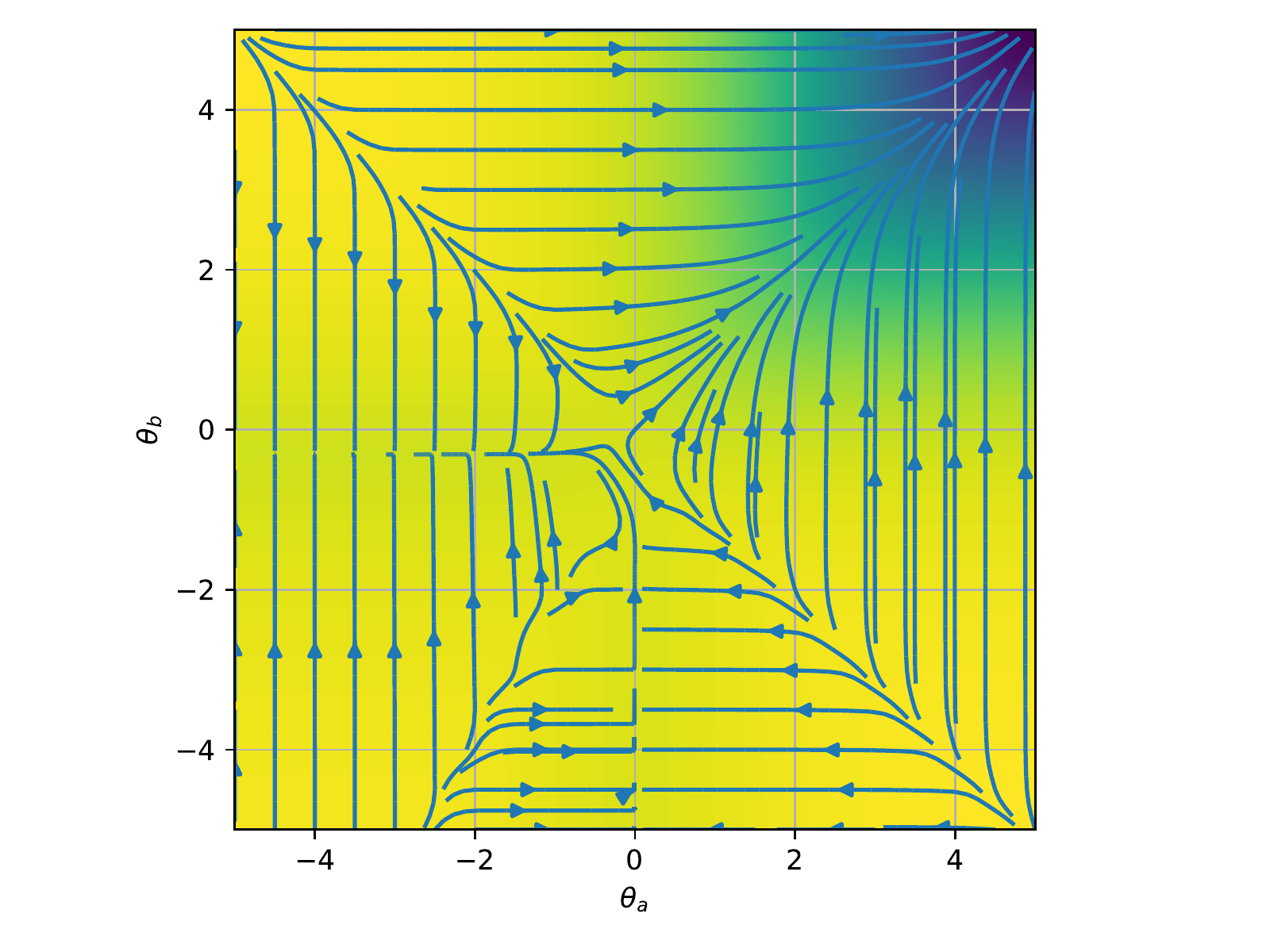}
\vspace{-6mm}
\caption{$L_{\text{generative}}$ for generative model (\Cref{sec:fullsum:app:generative-sum}).%
}
\label{fig:fullsum:app:loss-map-generative-sum}
\end{subfigure}%
\vspace{-2mm}
\caption{Different loss functions,
plotted over the model parameter space,
for \Cref{def:fullsum:example1} ($\exre$) \& \Cref{fullsum:example1:input} ($x_1^T$).
A darker color represents a lower loss value.
We also plot the negative gradient map,
such that we can see for every possible parameter setting,
where gradient descent leads to.
A uniform distribution initialization starts in the center with $\theta_\charsyma = \theta_\charblank = 0$.
We can also see local optima.
All models with parameters in the left upper area have peaky behavior.
The constructed optimal solution is in the right upper corner.}
\end{figure*}

\begin{theorem}\label{fullsum:ffnn:becomes-peaky}
\label{sec:fullsum:app:discriminative-sum}
Consider the FFNN model uniformly initialized, e.g.~$W = 0$,
and \Cref{fullsum:example1:input} ($\exre$) with $n\ge4$, i.e.~$T\ge 16$,
i.e.~the dominant label is $\genblank = \charblank$.
When trained with $L$ with gradient descent,
the model converges to peaky behavior,
which is a suboptimal local optima,
and yields 100\% error rate.

\begin{proof}  
We can reparameterize the \gls{ffnn} by the very generic model-free setting:
\begin{align*}
p_t(s|x_t{=}x_\charsyma) & :=\softmax((\theta_\charsyma, -\theta_\charsyma)) [s] , \\
p_t(s|x_t{=}x_\charblank) & := \softmax((-\theta_\charblank, \theta_\charblank)) [s] .
\end{align*}
This can parameterize any possible discriminative distribution (if we allow $\infty$ as well),
and specifically exactly the same probability distributions as the \gls{ffnn}.
In this parameterization, we have 2 scalar parameters $\theta_\charsyma,\theta_\charblank$,
i.e.~our parameter space is in $\R^2$. 
We get the initial uniform distribution with $\theta_\charsyma = \theta_\charblank = 0$.
We visualize the loss function over the parameters
in \Cref{fig:fullsum:app:loss-map-discriminative-sum}.
There are two global optima for this loss under the parameters,
which are $\theta_\charsyma = \theta_\charblank = \infty$
and $\theta_\charsyma = -\theta_\charblank = \infty$.
In the first case, the discriminative model would output the label $\charsyma$ at input $x_\charsyma$,
and label $\charblank$ at input $x_\charblank$.
In the second case, the model would always output label $\charsyma$.
From the figure, we can see that there is a local optima in the region
$\theta_\charblank > 0, \theta_\charsyma < 0$,
and this is the local optima which we reach
when we start in $\theta_\charsyma = \theta_\charblank = 0$.
For all parameters in this region $\theta_\charblank > 0, \theta_\charsyma < 0$,
all peaky alignments have higher scores than all other alignments,
i.e.~the model always has peaky behavior.
%
%
Decoding with this FFNN with peaky behavior
always yields $\argmax_s p_t(s \mid x_1^T) = \genblank = \charblank$,
i.e.~the model has 100\% error rate.

We can explicitly calculate the gradients
by \Cref{rem:q-by-counting} and SymPy.
For the (``blank") frames $t$ with $x_\charblank = \left( \begin{smallmatrix}0 \\ 1\end{smallmatrix} \right)$,
we get
\[ \E_{t, x_t = x_\charblank}[ q_t(\genblank|x_1^T, W_0) ] = \frac{19 n^2 - 1}{6 n (4n + 1)} > 74\% \]
and for the (``label") frames $t$ with $x_a = \left( \begin{smallmatrix}1 \\ 0\end{smallmatrix} \right)$,
we get
\[ \E_{t, x_t = x_\charsyma}[ q_t(\genblank|x_1^T, W_0) ] = \frac{13 n^2 - 1}{6 n (4n + 1)} > 50\% . \]

We see from the figure
that we can never escape that local minima, 
because the gradients on the lines $\theta_\charsyma = 0, \theta_\charblank > 0$
and $\theta_\charsyma < 0, \theta_\charblank = 0$ 
points towards
the same region $\theta_\charblank > 0, \theta_\charsyma < 0$,
which means that gradient descent can not escape from this region.

Case 1, $\theta_\charsyma < 0, \theta_\charblank = 0$:
Define
\begin{align*}
C_\charsyma(s_1^T) & := \left|\Set{t}{s_{t} = \genblank, x_{t} = x_\charsyma}\right| \in \Set{0, \dots, 2n}, \\
p_\charsyma & := \softmax((\theta_\charsyma,-\theta_\charsyma))[\charblank] > 0.5, \\
p'_\charsyma & := \frac{p_\charsyma}{1 - p_\charsyma} > 1 .
\end{align*}
Then
\begin{align*}
 & \E_{t, x_t = x_\charblank}[ q_t(s|x_1^T, W_0) ] \\
&= \frac{1}{p(y_1^N|x_1^T)}
\frac{1}{2n}
\sum_{\substack{t, \\ x_t = x_\charblank}}
\sum_{c=0}^{2n}
\sum_{\substack{s_1^T : y_1^N,\\ s_t = s, \\ C_\charsyma(s_1^T) = c}}
{0.5}^{2n}
{p_\charsyma}^{c}
(1 - p_\charsyma)^{2n - c} \\
&= \frac{1}{p(y_1^N|x_1^T)}
\frac{1}{2n}
\left( 0.5 (1 - p_\charsyma) \right)^{2n}
\sum_{\substack{t, \\ x_t = x_\charblank}}
\sum_{c=0}^{2n}
\sum_{\substack{s_1^T : y_1^N,\\ s_t = s, \\ C_\charsyma(s_1^T) = c}}
{p'_\charsyma}^{c} \\
&= \frac{1}{p(y_1^N|x_1^T)}
\frac{1}{2n}
\left( 0.5 (1 - p_\charsyma) \right)^{2n}
\sum_{c=0}^{2n}
{p'_\charsyma}^{c}
\sum_{\substack{t, \\ x_t = x_\charblank}}
\sum_{\substack{s_1^T : y_1^N,\\ s_t = s, \\ C_\charsyma(s_1^T) = c}}
1 .
\end{align*}
Now we are back at counting.
Define
\begin{align*}
\mathcal{C}_\charsyma(s,c) & :=
\sum_{\substack{t, \\ x_t = x_\charblank}}
\sum_{\substack{s_1^T : y_1^N,\\ s_t = s, \\ C_\charsyma(s_1^T) = c}}
1 , \\
\Delta \mathcal{C}_\charsyma(c) & := \mathcal{C}_\charsyma(s{=}\charblank,c) - \mathcal{C}_\charsyma(s{=}\charsyma,c) .
\end{align*}
If $\sum_{c=0}^{2n} \Delta \mathcal{C}_\charsyma(c) {p'_\charsyma}^{c} > 0$,
we have shown that
$\E_{t, x_t{=}x_\charblank}[ q_t(s{=}\charblank | x_1^T, W_0) ] > \E_{t, x_t{=}x_\charblank}[ q_t(s{=}\charsyma | x_1^T, W_0) ]$.
Via SymPy, 
we can calculate that
\begin{align*}
\Delta \mathcal{C}_\charsyma(c)
& =
\begin{cases}
0, & c = 0, \\
4 n (n^2 - 1) \frac{1}{3}, & c = 2n, \\
2 n (c + n), & 0 < c < 2n . \\
\end{cases}
\end{align*}
Given that we have $n \ge 4$,
we get $\Delta \mathcal{C}_\charsyma(c) > 0$ for all $c > 0$,
and thus $\sum_{c=0}^{2n} \Delta \mathcal{C}_\charsyma(c) {p'_\charsyma}^{c} > 0$.
I.e.~gradient descent will increase $\theta_\charblank$,
i.e.~increase $p(s{=}\charblank | x{=}x_\charblank)$.

Case 2, $\theta_\charsyma = 0, \theta_\charblank > 0$:
Analogous to the other case, we define
\begin{align*}
C_\charblank(s_1^T) & := \left|\Set{t}{s_{t} = \charblank, x_{t} = x_\charblank}\right| \in \Set{0, \dots, 2n}, \\
p_\charblank & := \softmax((-\theta_\charblank,\theta_\charblank))[\charblank] > 0.5, \\
p'_\charblank & := \frac{p_\charblank}{1 - p_\charblank} > 1 .
\end{align*}
Then we get
\begin{align*}
& \E_{t, x_t {=} x_\charsyma}[ q_t(s|x_1^T, W_0) ] \\
&= \frac{1}{p(y_1^N|x_1^T)}
\frac{1}{2n}
\left( 0.5 (1 - p_\charblank) \right)^{2n}
\sum_{c=0}^{2n}
{p'_\charblank}^{c}
\sum_{\substack{t, \\ x_t = x_\charsyma}}
\sum_{\substack{s_1^T : y_1^N,\\ s_t = s, \\ C_\charblank(s_1^T) = c}}
1 .
\end{align*}
Now we are back at counting.
Define
\begin{align*}
\mathcal{C}_\charblank(s,c) & :=
\sum_{\substack{t, \\ x_t = x_\charsyma}}
\sum_{\substack{s_1^T : y_1^N,\\ s_t = s, \\ C_\charblank(s_1^T) = c}}
1 , \\
\Delta \mathcal{C}_\charblank(c) & := \mathcal{C}_\charblank(s{=}\charblank,c) - \mathcal{C}_\charblank(s{=}\charsyma,c) .
\end{align*}
Via SymPy, 
we can calculate that
\begin{align*}
\Delta \mathcal{C}_\charblank(c)
& = \begin{cases}
2n (2n^2 - 3n - 2) \frac{1}{3}, & c = 2n, \\
4n (n - 1), & c = 2n - 1, \\
2n (3c - 4n + 1), & n \le c < 2n - 1, \\
-2n^2, & c = n - 1, \\
-2n (c + 1), & 0 \le c < n - 1 , \\
\end{cases} \\
\sum_{c=0}^{2n} \Delta \mathcal{C}_\charblank(c) &= 4n (n^2 - 3n - 1) \frac{1}{3} > 0.
\numberthis \label{eq:sum_c_b_gt_0}
\end{align*}
Define $c^* := \frac{4n-1}{3}$.
We see that
\begin{align*}
\Delta \mathcal{C}_\charblank(c) &= 0, \quad c = c^*, \\
\Delta \mathcal{C}_\charblank(c) &> 0, \quad \forall c > c^*, \\
\Delta \mathcal{C}_\charblank(c) &< 0, \quad \forall c < c^* .
\end{align*}
Now choose any $\tilde{c} \in \N$ with
\begin{align*}
\Delta \mathcal{C}_\charblank(c) &\ge 0, \quad \forall c \ge \tilde{c}, \\
\Delta \mathcal{C}_\charblank(c) &\le 0, \quad \forall c \le \tilde{c} - 1 .
\end{align*}
Via \Cref{eq:sum_c_b_gt_0}, we know that
\[ \sum_{c=\tilde{c}}^{2n} \Delta \mathcal{C}_\charblank(c) > - \sum_{c=0}^{\tilde{c}-1} \Delta \mathcal{C}_\charblank(c) . \]
And we get
\begin{align*}
- \sum_{c=0}^{\tilde{c}-1} \Delta\mathcal{C}_\charblank(c) {p'_\charblank}^{c}
\le& \left(- \sum_{c=0}^{\tilde{c}-1} \Delta\mathcal{C}_\charblank(c) \right) {p'_\charblank}^{ \tilde{c} } \\
<& \left( \sum_{c=\tilde{c}}^{2n} \Delta \mathcal{C}_\charblank(c) \right) {p'_\charblank}^{ \tilde{c} }
\le \sum_{c=\tilde{c}}^{2n} \Delta \mathcal{C}_\charblank(c) {p'_\charblank}^{c} ,
\end{align*}
and thus
\[ \sum_{c=0}^{2n}  \Delta\mathcal{C}_\charblank(c) {p'_\charblank}^{c} > 0 . \]
As before, it follows that
\[ \E_{t, x_t {=} x_\charsyma}[ q_t(s{=}\charblank | x_1^T, W_0) ] > \E_{t, x_t = x_\charsyma}[ q_t(s{=}\charsyma | x_1^T, W_0) ] . \]
I.e.~gradient descent will decrease $\theta_\charsyma$,
i.e.~increase $p(s{=}\charblank|x{=}x_\charsyma)$.
This results in peaky behavior,
and in 100\% error rate.
\end{proof}
\end{theorem}

These observation were shown for this specific constructed simple example,
however it can be argued that a similar behavior will usually be observed in other cases.
To emphasize:
\emph{A uniformly initialized \gls{ffnn}
trained with gradient descent on the \gls{ctc} loss does not converge to a global optimum,
but to a local optima with peaky behavior and 100\% error rate.}
The global optima of $L$ and all parameters close to that
have a perfect 0\% error rate without peaky behavior.
So this is mostly a problem of the gradient,
which tends towards peaky behavior,
and the model is too weak to be able to handle peaky behavior.


\begin{simulation}
We use \Cref{def:fullsum:example1} ($\exre$) and \Cref{fullsum:example1:input}, and $T=16$.
We see that the converged FFNN model has peaky behavior,
more specifically $p_t(s{=}\charblank | x_1^T) > 88\%$ for all $t$,
and 100\% error rate.
\end{simulation}

\begin{remark}
\label{rem:fullsum:bias}
\label{rem:fullsum:nobias}
If there is a global bias 
like it is usually the case for neural networks before the $\softmax$,
it will reinforce the convergence to peaky behavior 
because the gradient to the bias will be as in \Cref{th:fullsum:biasmodel}.
\end{remark}

The FFNN converges towards peaky behavior
but cannot learn the peaky alignment
because it has only local context.
We can argue that a more powerful model with global context
can always learn such alignment.
As a synthetic experiment, we introduce the \emph{memory model},
which has perfect memory.
This is the equivalent behavior of any model which can perfectly overfit.
I.e.~by construction this is the most powerful model possible.
This model is independent from the input $x$.


\begin{mydef}[Memory model]\label{def:fullsum:memmodel}
Define the model $\mathcal{M}^M$ with perfect memory as
\[
p_t(s|x_1^T,\mathcal{M}^M) := \softmax(M[t])[s]
\]
for $\theta = M \in \R^{T \times S}$.
\end{mydef}

\begin{simulation}
For \Cref{def:fullsum:example1} ($\exre$), $T=100$,
the memory model 
starting from uniform initialization
trained with $L$ with gradient descent
converges to peaky behavior
with $p_t(s{=}\charblank|x) > 93\% \ \forall t$, 
i.e.~100\% error rate.
\end{simulation}

\section{Role of the $\blank$ Label}

Recall \Cref{rem:fullsum:dom-label-by-topo}.
The $\blank$ label plays a special role in the \gls{ctc} topology.
We have seen that it is the dominant label,
and models tend to become peaky w.r.t.~the $\blank$ label.
It is important to point out that $\blank$ can occur anywhere in the alignment,
between all other labels.
In the common HMM topology in speech recognition,
there is no $\blank$, but $\sil$ instead.
Simply by counting, $\sil$ is also the dominant label.
Note that in the Wav2Letter \cite{collobert2016wav2letter}
label topology,
we have a special repetition label if a label is supposed to repeat
on the target side.
The Chain model \cite{povey2016purely}
label topology has two states per phoneme,
where the second optional looping label
is interpret as blank -- however, it is not shared,
and thus not dominant.
Both Wav2Letter and Chain have a dominant $\sil$ label as well.
However, they use other training criteria
which do not necessarily lead to peaky behavior.

CTC trained with dominant $\sil$ label results in peaky models.
However, the label topology usually allows that
$\sil$ can only occur before or after whole words, not within words,
where a word consists of multiple phonemes.
We will argue that this label topology
is suboptimal for a loss like CTC with peaky behavior.
We will construct an even simpler example for the further demonstration.

\begin{figure*}
\centering
\begin{subfigure}[b]{0.33\linewidth}
\centering
\includegraphics[width=\linewidth]{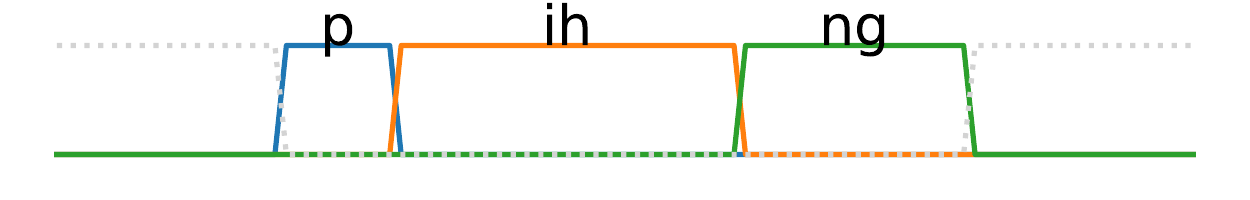}
\vspace{-6mm}
\caption{Optimal with $\sil$.}
\label{fig:fullsum:ex-ping-opt-sil}
\end{subfigure}%
\begin{subfigure}[b]{0.33\linewidth}
\includegraphics[width=\linewidth]{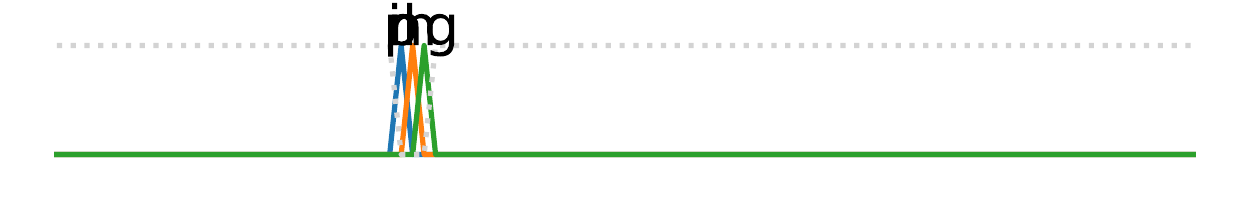}
\vspace{-6mm}
\caption{Peaky with $\sil$.}
\label{fig:fullsum:ex-ping-peaky-sil}
\end{subfigure}%
\begin{subfigure}[b]{0.33\linewidth}
\centering
\includegraphics[width=\linewidth]{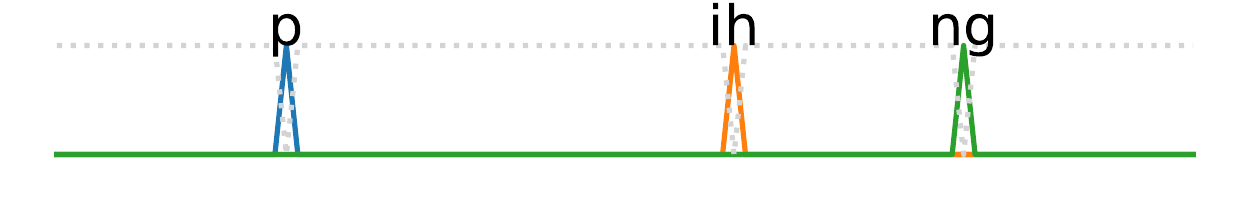}
\vspace{-6mm}
\caption{Peaky with $\blank$.}
\label{fig:fullsum:ex-ping-peaky-blank}
\end{subfigure}
\vspace{-2mm}
\caption{Like in \Cref{fig:ctc-full-sum-peaky-out-2},
colors represents the posterior outputs for the labels ``p", ``ih" and ``ng".
The dotted gray output represents $\sil$ or $\blank$.
The posteriors are constructed such that they represent a Viterbi alignment.}
\label{fig:fullsum:ex-ping-peaky-blank-vs-sil}
\end{figure*}

\begin{example}\label{ex:fullsum:ping}
Let us consider the single word ``ping"
which consists of the phoneme sequence ``p ih ng".
With the \gls{ctc} topology, $\blank$ is allowed anywhere.
With the standard \gls{hmm} topology,
$\sil$ is allowed only before ``p" and after ``ng".
We will demonstrate that this restriction is suboptimal
together with peaky behavior which results by the loss $L$,
and a $\blank$ label which can occur everywhere is better.
We further construct corresponding input features
\[ x_1^T = \left(
\smash[b]{\underbrace{\begin{matrix}0 \\ 0 \\ 0 \\ 1\end{matrix} \cdots \begin{matrix}0 \\ 0 \\ 0 \\ 1\end{matrix}}_{20 \times}}\;
\smash[b]{\underbrace{\begin{matrix}1 \\ 0 \\ 0 \\ 0\end{matrix} \cdots \begin{matrix}1 \\ 0 \\ 0 \\ 0\end{matrix}}_{10 \times}}\;
\smash[b]{\underbrace{\begin{matrix}0 \\ 1 \\ 0 \\ 0\end{matrix} \cdots \begin{matrix}0 \\ 1 \\ 0 \\ 0\end{matrix}}_{30 \times}}\;
\smash[b]{\underbrace{\begin{matrix}0 \\ 0 \\ 1 \\ 0\end{matrix} \cdots \begin{matrix}0 \\ 0 \\ 1 \\ 0\end{matrix}}_{20 \times}}\;
\smash[b]{\underbrace{\begin{matrix}0 \\ 0 \\ 0 \\ 1\end{matrix} \cdots \begin{matrix}0 \\ 0 \\ 0 \\ 1\end{matrix}}_{20 \times}}
\right)
\vphantom{\underbrace{\begin{matrix}0 \\ 0 \\ 0 \\ 1\end{matrix} \cdots \begin{matrix}0 \\ 0 \\ 0 \\ 1\end{matrix}}_{20 \times}}
\]
with $T = 100$.
This example is constructed such that the time accurate (optimal) alignment is
$s_1^T = (\charblank^{20},\charsyma^{10}, \charsymb^{30}, \charsymc^{20}, \charblank^{20})$,
and an optimal model is
\[ p(s | x) := \begin{cases}1, & x = x_s\\ 0, & \text{else}\end{cases} \]
with $x_s$ accordingly.
The posteriors of this model are visualized in \Cref{fig:fullsum:ex-ping-opt-sil}.
\end{example}

\begin{remark}
\label{thm:fullsum:ffnn-train-difficulty}
Consider the case for the \gls{hmm} topology with $\sil$,
i.e.~we allow all alignments matching the regular expression
$(\charblank^*, \charsyma^+, \charsymb^+, \charsymc^+, \charblank^*)$.
Peaky behavior results in alignments of the form
$(\charblank^+, \charsyma, \charsymb, \charsymc, \charblank^+)$.
Optimal posteriors of this alignment are visualized in \Cref{fig:fullsum:ex-ping-peaky-sil}.
%
With the \gls{ctc} topology, i.e~with $\blank$,
we allow all alignments matching the regular expression
$(\charblank^*, \charsyma^+, \charblank^*, \charsymb^+, \charblank^*, \charsymc^+, \charblank^*)$.
We get the peaky behavior with alignments of the form
$(\charblank^+, \charsyma, \charblank^+, \charsymb, \charblank^+, \charsymc, \charblank^+)$.
Optimal posteriors of this alignment are visualized in \Cref{fig:fullsum:ex-ping-peaky-blank}.
%
Comparing both possible posteriors and alignments,
we see that the \gls{hmm} topology is much more restricted,
and peaky behavior compresses a whole word as short as possible.
This is clearly suboptimal,
as it was also experimentally observed \cite{zeyer2017:ctc}.
\end{remark}


\begin{remark}
This implies that a $\blank$ label can help in general for CTC training.
This is also true if the modeling is performed on phone-level,
and in fact this seems to work well in practice
\cite{sak2015ctc,miao2016empiricalctc}.
\end{remark}


\section{Role of the Ratio $T/N$}
\label{sec:fullsum:seq-len-ratio}

From
\Cref{cor:fullsum:ex1:countblankframes}
we can see that the peaky behavior
is amplified the higher the ratio $\frac{T}{N}$
purely due to the label topology.

\begin{figure}
\centering
\includegraphics[width=1.\linewidth]{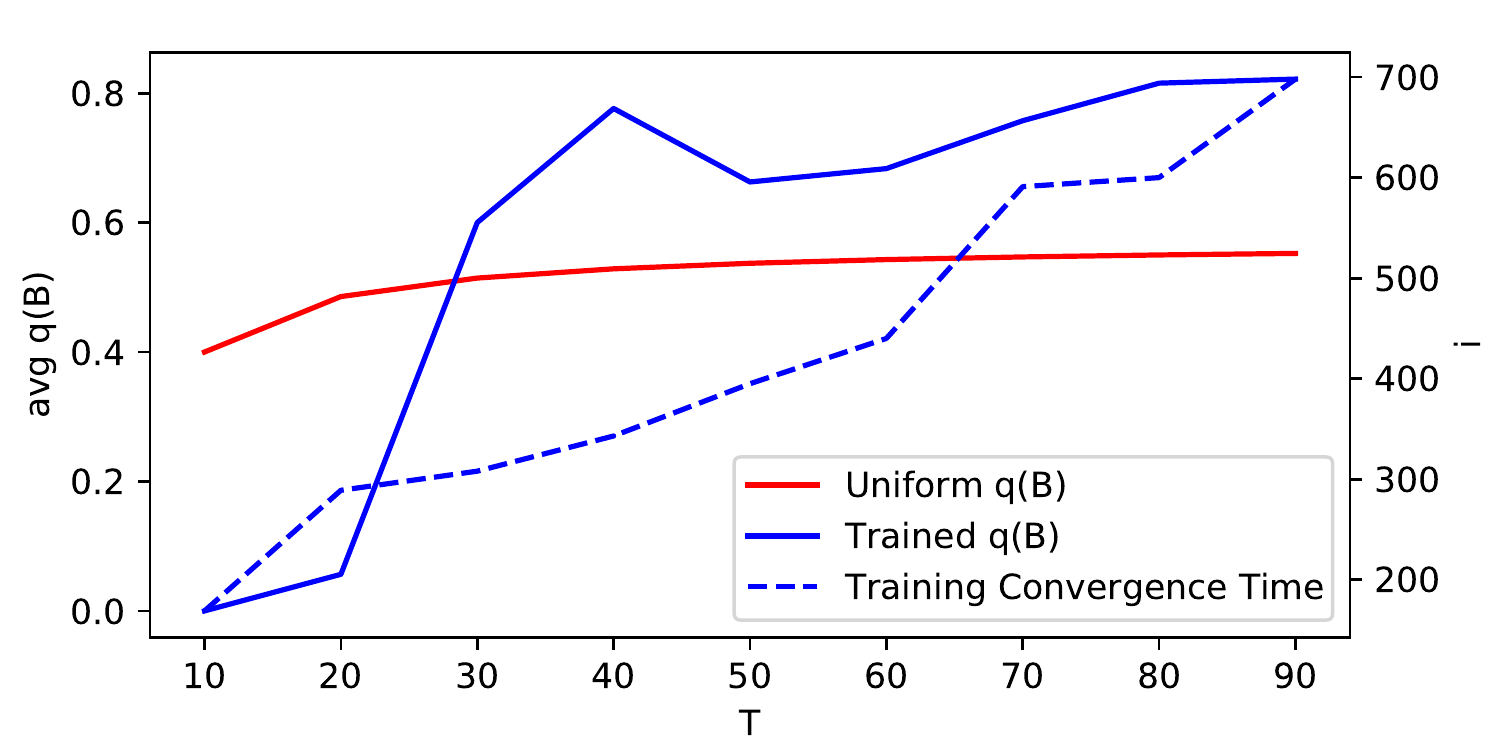}
\vspace{-8mm}
\caption{This uses the target sequence $\charsyma\charsymb\charsymc$ with $N=3$,
and $x_1^T$ constructed as in \Cref{ex:fullsum:ping}
but downscaled accordingly.
We plot $\frac{1}{T} \sum_t q_t(\charblank)$
for a uniform distribution $p$,
and of a CTC-trained LSTM model.
We also plot the convergence time,
which measures the number of steps $i$ until $L < 1$,
i.e.~lower is better.}
\label{fig:fullsum:ctc-ratio-t-n}
\end{figure}

\begin{simulation}
We use the same example from \Cref{ex:fullsum:ping}
for the target $y_1^N = \charsyma\charsymb\charsymc$ ($N=3$),
\Gls{ctc} label topology (including $\blank$)
and $x_1^T$ synthetically constructed for varying $T$,
where $p(x)$ stays uniform.
We want to study the effect of the ratio $\frac{T}{N}$
on the peaky behavior and convergence behavior.
We can measure the average $q(\charblank)$
for a uniform distribution $p$ to get the the initial gradient
due to the label topology and $\frac{T}{N}$.
We train a simple \gls{lstm} \cite{hochreiter1997lstm} model with \gls{ctc},
and measure the resulting average $q(\charblank)$
which shows how dominant $\charblank$ has become.
The model learns perfectly in all cases,
although with varying convergence speed.
We plot the results in
\Cref{fig:fullsum:ctc-ratio-t-n}.
For $T \le 20$ ($\frac{T}{N} < 7$),
we observe that the model does not converge
to peaky behavior,
while it tends to for larger $T$.
Also, we see that the convergence speed decreases
with increasing $T$,
which indicates that a high $\frac{T}{N}$ ratio is harder to learn.
\end{simulation}

\section{Avoiding Peaky Behavior by Other Losses}
\label{sec:fullsum:avoid-peaky-behavior}

For good error rate performance,
avoiding peaky behavior might not be needed.
However,
peaky behavior can be problematic in certain cases,
e.g.~when an application requires to not use the $\blank$ label
(e.g.~for time accurate phoneme or word boundaries in the alignment),
or for the usage of local-context models like \glspl{ffnn},
as we have shown.
%

We extend the training criterion by a label prior
and show that this does not lead to peaky behavior.
We will demonstrate that this solves the convergence issues for \glspl{ffnn}.
This loss is originated in the \gls{hybrid-hmm} model case
\cite{bourlard1989hybridhmm,franzini1990hybridhmm}.
where the generative acoustic model integrates a discriminative \gls{nn}
by
\[ p(x_1^T|s_1^T) \propto \frac{p(s_1^T | x_1^T)}{p(s_1^T)} . \]
The difference here to the usual \gls{ctc} model is the label prior model $p(s_1^T)$ in the denominator.
It can even be useful for decoding with \gls{ctc} models \cite{miao2015eesen}.
We usually simplify the \emph{prior model} $p(s_1^T)$ as
\[ p(s_1^T | \mathcal{M}_{\text{prior}}) = \prod_t p(s_t | \mathcal{M}_{\text{prior}}) . \]
Other prior variants are possible \cite{kanda2016mapctc}. 

\begin{mydef}
\label{def:fullsum:lossprior}
Define the \emph{hybrid model loss} as 
\begin{align*}
L_{\text{hybrid}} &:= -\log \sum_{s_1^T:y_1^N} \frac{p(s_1^T|x_1^T, \mathcal{M})}{p(s_1^T | \mathcal{M}_{\text{prior}})} \\
&\phantom{:}= -\log \sum_{s_1^T:y_1^N} \prod_t \frac{p_t(s_t|x_1^T, \mathcal{M})}{p(s_t | \mathcal{M}_{\text{prior}})} .
\end{align*}
\end{mydef}
$L_{\text{hybrid}}$ was used in \cite{haffner1993connectionistmmi,zeyer2017:ctc}.

There are multiple options how to estimate the prior $p(s)$.
\begin{remark}
Having a prior model $p(s|\mathcal{M}_\text{prior}) = \softmax(b_\text{prior})[s]$
as a separate model with its own parameters,
and trained jointly with the posterior model $p_t(s|x_1^T,\mathcal{M}_\text{posterior})$
will lead exactly to an inverse prior estimation.
I.e.\ consider that a label $\genblank$ maximizes the posterior model $p(s_1^T|x_1^T)$,
i.e.\ it would occur most often in Viterbi alignments (disregarding the prior model).
Then, the prior model would be optimal for minimizing $L_\text{hybrid}$
when it \emph{minimizes} $p(\genblank)$, i.e.\ $p(\genblank) < p(s)$.
This is counter intuitive and does not reflect what the prior model should represent.
Also, it would only reinforce the peaky behavior.
\end{remark}
Given this remark, it becomes clear that $p(s)$ should be estimated
based on the posterior model in some way.

\begin{mydef}
\label{def:fullsum:softmaxprior}
\label{rem:fullsum:softmaxprior}
Letting the prior model $p(s)$ be estimated as the expected output of the posterior model,
which we also call \emph{softmax prior}, cf.\ \cite{manohar2015semi},
i.e.\
\[ p(s) := \frac{1}{T} \sum_t p_t(s|x_1^T) . \]
\end{mydef}

\begin{remark}
\label{sec:fullsum:app:discriminative-sum-with-prior}
%
%
Just as in \Cref{sec:fullsum:app:discriminative-sum}, for the same example,
with the same parameterization,
we plot the loss function $L_{\text{hybrid}}$ in \Cref{fig:fullsum:app:loss-map-discriminative-sum-with-prior}.
We can see that there is only a single global optimum at $\theta_a = \theta_b = \infty$,
and also that we reach that global optimum at a uniform distribution initialization ($\theta_a = \theta_b = 0$).
%
\end{remark}




\begin{simulation}
We use the FFNN model (\Cref{def:fullsum:ffnn})
with softmax prior (\Cref{def:fullsum:softmaxprior}),
and
\Cref{def:fullsum:example1} ($\exre$),
$x_1^T$ as in \Cref{fullsum:example1:input}.
We can see that
training with the loss $L_{\text{hybrid}}$
will not get peaky behavior.
The model converges to the
time accurate (optimal) alignment.
I.e.~it converges towards
$W = \left( \begin{smallmatrix} \infty & 0 \\ 0 & \infty \end{smallmatrix} \right)$
with 0\% error rate.
%
\end{simulation}


\begin{remark}
\label{sec:fullsum:app:discriminative-sum-with-prior-sg}
The common training of \gls{hybrid-hmm} models would keep the prior model $p(s)$ fixed
while updating the posterior model $p(s|x)$.
In our formulation, that is equivalent by defining
\[ p(s | \mathcal{M}_{\text{prior-sg}}) := \operatorname{stop-gradient}\left( \frac{1}{T} \sum_t p_t(s|x_1^T,\mathcal{M}) \right) , \]
where $\operatorname{stop-gradient}$ is the identity function,
but the gradient is defined as zero.
In that case, the gradient of $L_{\text{hybrid}}$ will look different.
We can see the effect in \Cref{fig:fullsum:app:loss-map-discriminative-sum-with-prior-sg}.
We observe a slightly different behavior of the gradient map.
However, starting with uniform distribution initialization ($\theta_a = \theta_b = 0$)
will converge to the same global optimum.
\end{remark}


\begin{remark}
\label{rem:fullsum:genhybridpeaky}
When the prior is kept fixed but the dominance of $\genblank$ is strong enough,
this still can lead to peaky behavior.
Alternatively, if the prior is too strong or not well estimated,
this can result in the dominance of another label $\hat{s} \neq \genblank$
in $\frac{p(\hat{s} \mid x)}{p(\hat{s})}$,
and can get peaky behavior where this other label $\hat{s} \neq \genblank$ dominates.
We observed this behavior in some cases experimentally,
where we used an online moving average of $p(s\mid x)$ for the prior.
This online moving average estimation can be unstable,
esp.~in early stages of training.
\end{remark}

\begin{remark}
%
A stable recipe is to estimate the prior on the whole training data
as in \Cref{def:fullsum:softmaxprior},
then to calculate the soft alignment $q_t$ for the whole training data,
and then to update the posterior model while keeping the soft alignments fixed.
An approximation of using the soft alignment are hard Viterbi alignments.
This is very similar to the standard training procedure for \glspl{hybrid-hmm}
with framewise \gls{ce}.
\end{remark}

The peaky behavior was only observed for discriminative models,
while similar training criteria have been used for generative models.
We now study the convergence behavior and peaky behavior of generative models.
We use a simple generative model (without transition probabilities)
\begin{align*}
p(x_1^T|y_1^N,\mathcal{M})
& \propto  \sum_{s_1^T:y_1^N} p(x_1^T|s_1^T,\mathcal{M})  \\
& =  \sum_{s_1^T:y_1^N}\prod_t p(x_t|s_t,\mathcal{M}) .
\end{align*}

\begin{mydef}[Loss for generative model]
Define the loss
\begin{align*}
L_{\text{generative}} & :=
-\log \sum_{s_1^T:y_1^N} p(x_1^T|s_1^T,\mathcal{M}) \\
& \phantom{:}= -\log \sum_{s_1^T:y_1^N} \prod_t p(x_t|s_t,\mathcal{M}) .
\end{align*}
\end{mydef}


We follow a similar construction as for the FFNN (\Cref{sec:fullsum:app:discriminative-sum})
for \Cref{def:fullsum:example1} ($\exre$),
$x_1^T$ as in \Cref{fullsum:example1:input}.

\begin{mydef}[Generative model]
\label{def:fullsum:generative-model}
For $S = \Set{\charblank, \charsyma}$, $x \in \Set{x_\charblank, x_\charsyma}$,
we define
\begin{align*}
p(x|s{=}\charsyma) &= \softmax((\theta_\charsyma, -\theta_\charsyma)) \left[ \begin{cases} 1, & x = x_\charsyma \\ 2, & x = x_\charblank \end{cases} \right] , \\
p(x|s{=}\charblank) &= \softmax((-\theta_\charblank, \theta_\charblank)) \left[ \begin{cases} 1, & x = x_\charsyma \\ 2, & x = x_\charblank \end{cases}\right] .
\end{align*}
\end{mydef}

\begin{remark}
\label{sec:fullsum:app:generative-sum}
We plot the loss $L_{\text{generative}}$ in \Cref{fig:fullsum:app:loss-map-generative-sum}.
We can see that there is only a single global optimum at $\theta_\charsyma = \theta_\charblank = \infty$
with $L=0$.
When we start with uniform distribution
($\theta_\charsyma = \theta_\charblank = 0$),
we reach that global optimum.
This global optimum
is the optimal non-peaky solution,
and the error rate becomes 0\%.
\end{remark}

\begin{remark}
We can reparameterize the model as 
\begin{align*}
p(x | s) := \begin{cases} \theta[s], & x = x_s \\ 1 - \theta[s], & x \ne x_s \end{cases} .
\end{align*}
For $\theta \equiv 1$, we get the unique global optimum with $L=0$,
and this global optimum has no peaky behavior.
We get our initial uniform distribution with $\theta_0 \equiv 0.5$.
%
We will get a similar gradient as in \Cref{eq:fullsum:L-grad},
however, this error signal is for the generative model $p(x_t|s_t)$.
\begin{align*}
\frac{\partial L_{\text{generative}}}{\partial \theta} &=
-\sum_{s,t} q_t(s) \frac{\partial}{\partial \theta} \log p(x_t|s, \theta) \\
&= -\sum_{s,x} \left(\sum_{t, x_t = x} q_t(s) \right) \frac{\partial}{\partial \theta} \log p(x|s, \theta) .
\end{align*}
Assume $\theta_i[s] \not\in \Set{0,1}$.
Then for any $s$,
\begin{align*}
\frac{\partial L_{\text{generative}}}{\partial \theta_i[s]}
= - \frac{\sum_{t, x_t = x_s} q_t(s,\theta_i)}{\theta_i[s]} + \frac{\sum_{t, x_t \ne x_s} q_t(s,\theta_i)}{1 - \theta_i[s]} .
\end{align*}
Following a similar calculation as in \Cref{fullsum:ffnn:becomes-peaky},
we can explicitly calculate that for $\theta_0 = 0.5$,
\begin{align*}
\frac{\sum_{t, x_t = x_\charblank} q_t(s{=}\charblank,\theta_0)}{\sum_t q_t(s{=}\charblank,\theta_0)}
&= \frac{19 n^2 - 1}{32 n^2 - 2} > 59\% \\
\frac{\sum_{t, x_t = x_\charsyma} q_t(s{=}\charsyma,\theta_0)}{\sum_t q_t(s{=}\charsyma,\theta_0)}
&= \frac{11 n^2 + 6n + 1}{16 n^2 + 12 n + 2} \ge 60\% .
\end{align*}
Thus 
$\frac{\partial L_{\text{generative}}}{\partial \theta_0[s]} < 0$,
i.e.~$\theta_1[s] > \theta_0[s]$.
\end{remark}

\begin{simulation}
For $L_{\text{generative}}$,
for \Cref{def:fullsum:example1} ($\exre$),
$x_1^T$ as in \Cref{fullsum:example1:input}, $T=16$,
and the model parameterized as in \Cref{def:fullsum:generative-model},
we see that the model converges to the
global optimum with
time accurate (optimal) alignment,
i.e.~it does not get peaky behavior
and has 0\% error rate.
\end{simulation}

\section{Conclusions}

We contribute a formal analysis to discover the causes for peaky behavior.
We found this is a property of local convergence which tends towards peaky behavior
when starting from a uniform distribution.
This is due to the label topology and the dominance of one label such as $\blank$ or $\sil$.
We also explained
the role of the $\blank$ label, the role of the ratio $\frac{T}{N}$,
and the role of a label prior model
in \gls{ctc} and full-sum training.
We have shown that peaky behavior should be avoided without a $\blank$ label.
Even with the $\blank$ label and \gls{ctc} topology,
peaky behavior can be suboptimal,
as was demonstrated on a simple example with a simple \gls{ffnn}.
We extended the training criterion to handle and avoid the peaky behavior
by including a label prior.


\bibliography{peaky_behavior}
\bibliographystyle{icml2021}

\end{document}